\documentclass[10pt, a4paper]{article}

\usepackage[]{lrec-coling2024} 
\usepackage{amssymb}
\usepackage{amsmath}
\usepackage{booktabs}
\usepackage{multirow}
\title{A Semantic Mention Graph Augmented Model for Document-Level\\ \vspace*{.5\baselineskip}   Event Argument Extraction}

\name{Jian Zhang\textsuperscript{1}, Changlin Yang\textsuperscript{1}, Haiping Zhu\textsuperscript{1,2}\sthanks{~~Corresponding author.}, Qika Lin\textsuperscript{3}, Fangzhi Xu\textsuperscript{1}, Jun Liu\textsuperscript{1,2}} 

\address{\textsuperscript{1}School of Computer Science and Technology, Xi'an Jiaotong University \\
\textsuperscript{2}National Engineering Lab for Big Data Analytics, Xi'an, China\\
\textsuperscript{3}National University of Singapore \\
        \{zhangjian062422,yangchanglin,Leo981106\}@stu.xjtu.edu.cn, \\
        \{zhuhaiping,liukeen\}@xjtu.edu.cn,
        linqika@nus.edu.sg
        }

\abstract{
Document-level Event Argument Extraction (DEAE) aims to identify arguments and their specific roles from an unstructured document. The advanced approaches on DEAE utilize prompt-based methods to guide pre-trained language models (PLMs) in extracting arguments from input documents. They mainly concentrate on establishing relations between triggers and entity mentions within documents, leaving two unresolved problems: a) independent modeling of entity mentions; b) document-prompt isolation. To this end, we propose a semantic mention Graph Augmented Model (GAM) to address these two problems in this paper. Firstly, GAM constructs a semantic mention graph that captures relations within and between documents and prompts, encompassing co-existence, co-reference and co-type relations. Furthermore, we introduce an ensembled graph transformer module to address mentions and their three semantic relations effectively. Later, the graph-augmented encoder-decoder module incorporates the relation-specific graph into the input embedding of PLMs and optimizes the encoder section with topology information, enhancing the relations comprehensively. Extensive experiments on the RAMS and WikiEvents datasets demonstrate the effectiveness of our approach, surpassing baseline methods and achieving a new state-of-the-art performance.
 \\ \newline \Keywords{document-level event argument extraction, semantic mention graph, ensembled graph transformer, graph-augmented PLMs} }

\begin{document}

\maketitleabstract

\section{Introduction}\label{1}

\begin{figure}
\begin{center}
  \includegraphics[scale=0.5]{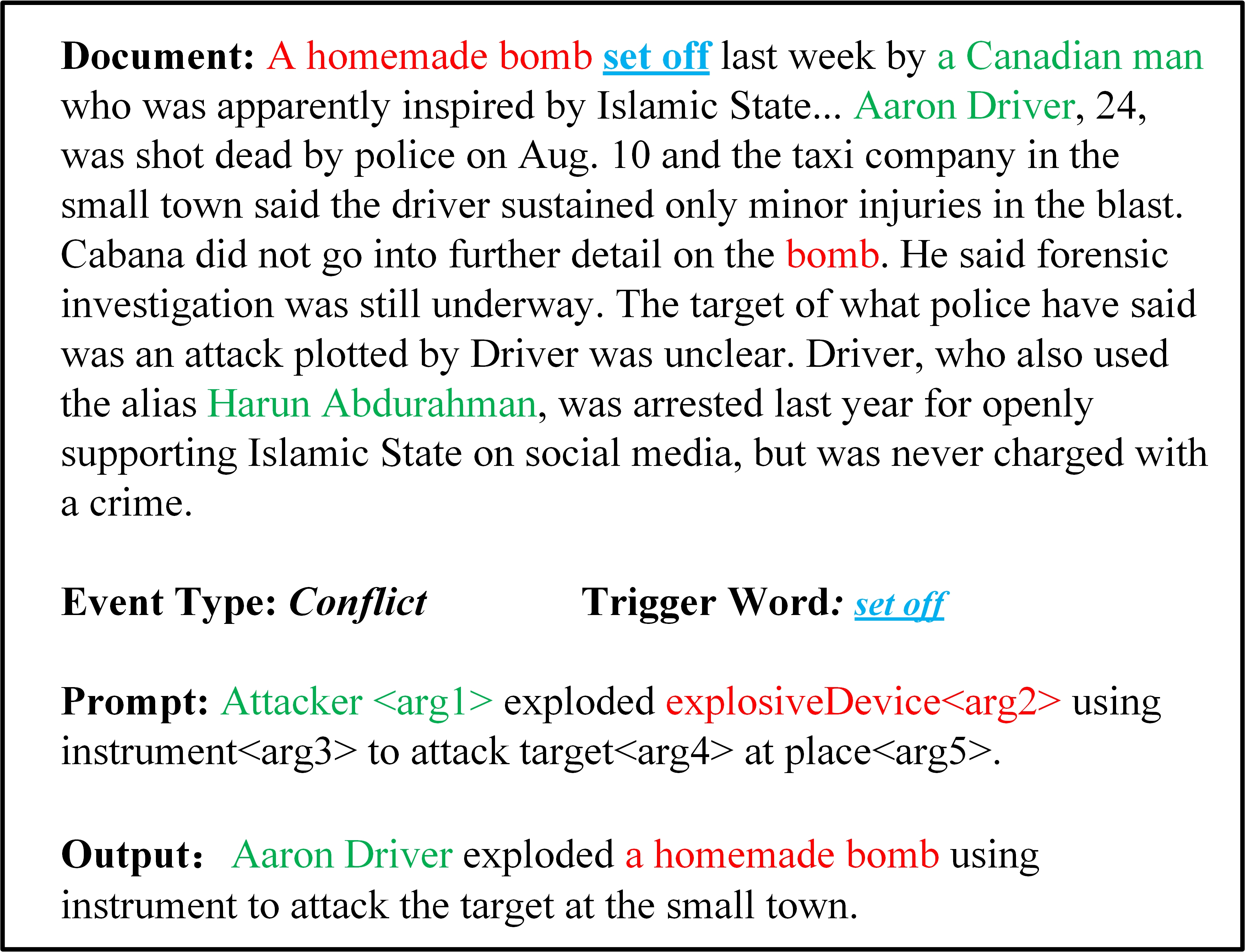}
  \caption{An illustration of DEAE including the relevance among entity mentions with the same color labeled in the document and the co-type relation between the prompt and 
 the document with the same color labeled.}
  \vspace{-0.6cm}
  \label{fig_dataset}
\end{center}
\end{figure}

Document-level Event Extraction (DEE) stands as an essential technology in the construction of event graphs~\cite{xu2021document} in the field of natural language processing (NLP)~\cite{hirschberg2015advances,hedderich-etal-2021-survey,bojun2023utility}. Within the realm of DEE, Document-level Event Argument Extraction (DEAE) plays a crucial role in transforming unstructured text into a structured event representation, thereby enabling support for various downstream tasks like recommendation systems~\cite{roy2022systematic}, dialogue systems~\cite{ni2023recent} and some reasoning applications~\cite{wang2023hugnlp}.
DEAE strives to extract all arguments from the entity mentions in a document and assign them specific roles with a given trigger word representing the event type. As depicted in Fig.~\ref{fig_dataset}, the trigger word is \textit{set off} and the task is to extract arguments of the predefined argument roles of the event type \textit{Conflict}, e.g., \textit{attacker} and \textit{explosiveDevice}.
In recent researches, significant strides have been made in DEAE thanks to the success of pre-trained language models (PLMs) and the prompt-tuning paradigm. An unfilled prompt $p$ is initialized by argument placeholders based on the event ontology\cite{li2021document}. For example, the prompt for \textit{Conflict} type in Fig.~\ref{fig_dataset} is ``\textit{Attacker} $\left \langle arg1 \right \rangle$ \textit{exploded explosiveDevice} $\left \langle arg2 \right \rangle$ \textit{using instrument} $\left \langle arg3 \right \rangle$ \textit{to attack target} $\left \langle arg4 \right \rangle$ \textit{at place} $\left \langle arg5 \right \rangle$''.
We define argument placeholders in the prompt as mask mentions, e.g., ``\textit{attacker} $\left \langle arg1 \right \rangle$''. The advanced approaches on DEAE utilize prompt-based methods to guide PLMs in extracting arguments from input documents. These studies on DEAE~\cite{lin2022cup,ma2022prompt,zeng2022ea2e} consider using different prompts to instruct PLMs, but there remains two unsolved problems: a) independent modeling of entity mentions; b) document-prompt isolation.

On one hand, the relevance among entity mentions within the document is crucial but frequently overlooked. These entity mentions share a clear and significant connection that demands careful consideration in DEAE. This relevance is universal and invaluable, enabling DEAE to grasp the completeness of events and the correlation structure within documents. Taking co-reference relations as an example, arguments appear in various forms across different sentences within the document, creating co-reference instances. As shown in Fig.~\ref{fig_dataset}, the entity mention \textit{Aaron Driver} appears multiple times in various forms of expressions (in green), such as \textit{a Canadian man} and \textit{Harun Abdurahman}, conveying an identical semantic meaning. The same phenomenon also exists in the co-existence relation among entity mentions, wherein co-existence relations denote the presence of entity mentions or masked mentions within the same sentence. Surprisingly, previous studies~\cite{du2020event,liu2021machine,wei2021trigger} often overlook this aspect, obscuring this vital correlation.

On the other hand, the document-prompt isolation is both valuable and underappreciated.  Generally, arguments should only be extracted from entity mentions of the same type in the appropriate context. The co-type relation between documents and prompts provides essential guidance for accurately determining the positions of arguments. In other words, the co-type relation refers to the same type attributes between masked mentions and entity mentions. As demonstrated in Fig.~\ref{fig_dataset}, the argument role \textit{attacker} in the prompt and the entity mention \textit{Aaron Driver} are of the same type, namely, \textit{PERSON}, indicating a co-type phenomenon. Previous studies~\cite{lin2022cup,ma2022prompt,zeng2022ea2e} ignore the document-prompt isolation problem, neglecting the co-type relation between documents and prompts when directly feeding them into PLMs.  

To this end, we propose a semantic mention Graph Augmented Model (GAM) to alleviate the above two problems in this paper.  Within the semantic mention graph, the semantics highlights the internal meaning of mentions and models this through the relations between mentions. To address the independent modeling of entity mentions, GAM considers the co-existence and co-reference relations among entity mentions. For the document-prompt isolation problem, the co-type relation between mask mentions and entity mentions are incorporated. Specifically, we first construct a semantic mention graph module to model these three semantic relations. It includes nodes representing entity mentions and mask mentions, connected by the aforementioned relations. For instance, nodes like \textit{Aaron Driver} and \textit{Harun Abdurahman} are connected by an edge labeled \textit{co-reference}. Then, the three types of relations are depicted in three adjacent matrices, which are aggregated into a fused attention bias. The node sequence and fused attention bias are fed into the ensembled graph transformer for encoding. Lastly, we integrate node embeddings into initial embeddings as input and employ the fused topology information as attention bias to boost the PLMs. The main contributions of our work are as follows:

\begin{itemize}
    \item{This research introduces a universal framework GAM\footnote{The code for the framework and the experimental data are stored in the repository: https://github.com/exoskeletonzj/gam.}, in which we construct a semantic mention graph incorporating three types of relations within and between the documents and the prompts initially. It is the first work in simultaneously addressing the independent modeling of entity mentions and document-prompt isolation as far as we know.}
    \item{We propose an ensembled graph transformer module and a graph-augmented encoder-decoder module to handle the three types of relations. The former is utilized to handle the mentions and their three semantic relations, while the latter integrates the relation-specific graph into the input embedding and optimizes the encoder section with topology information to enhance the performance of PLMs.}
    \item {Extensive experiments report that GAM achieves the new state-of-the-art performance on two benchmarks and further analysis validates the effectiveness of the different relations in semantic mention graph construction module, ensembled graph transformer module and graph-augmented encoder-decoder module in our model.}
\end{itemize}

\section{Related Works}

In this section, we introduce the current researches on DEAE, mainly consisting the sequence model and graph model for event extraction.

\subsection{DEAE Based on Sequence Model}

From the early stages, semantic role labeling (SRL) has been utilized for extracting event arguments in various studies~\cite{yang2018dcfee,zheng2019doc2edag,xu2021document,wang2023cok}. Some studies initially identify entities within the document and subsequently assign these entities specific argument roles. ~\citet{lin2020joint} began by identifying candidate entity mentions, followed by their assignment of specific roles through multi-label classification.

Later, certain studies have approached DEAE as a question-answering (QA) task. Methods~\cite{du2020event,liu2021machine} based on QA involve querying arguments by answering questions predefined through templates one by one, treating DEAE as a machine reading comprehension task. ~\citet{wei2021trigger} 
toke into account the implicit interactions among roles by imposing constraints on each other within the template. However, this method tends to lead to error accumulation.

\begin{figure*}[t]
	\large
	\centering
	\includegraphics[scale=0.5]{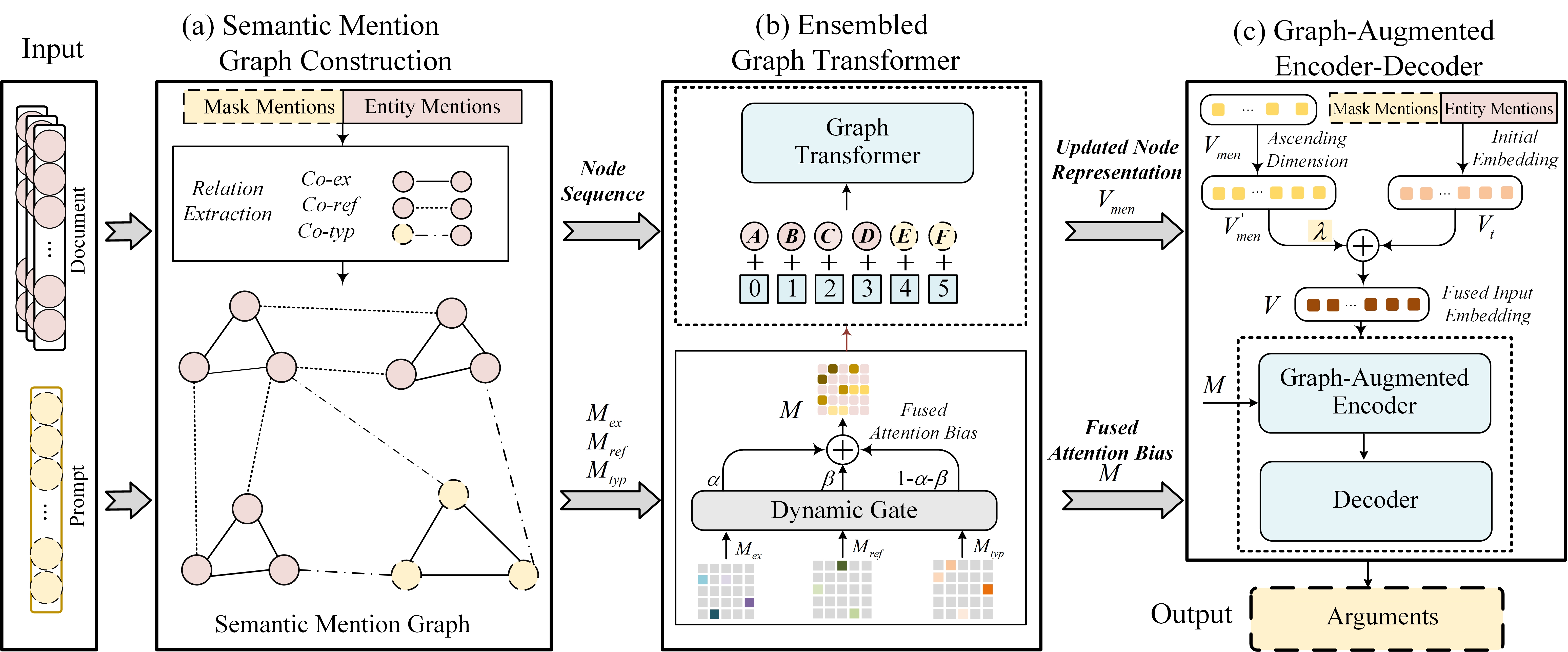}
	\caption{The architecture of GAM. The left part is an input example of the document and a corresponding prompt. The graph construction module (\textit{a}) constructs a semantic mention graph including co-existence, co-reference and co-type relations from entity mentions and mask mentions. The ensembled graph transformer module (\textit{b}) handles the text features combined with three semantic relations. Finally, the graph-augmented encoder-decoder module (\textit{c}) is utilized to conduct the feature fusion and predict the arguments.}
	\label{fig_model}
	\vspace{-0.3cm}
\end{figure*}

Alongside the emergence of sequence-to-sequence models, specifically generative PLMs like BART~\cite{lewis2020bart} and T5~\cite{raffel2020exploring}, generating all arguments in the sequence of target event has become possible. Some studies~\cite{li2021document,du2021template,lu2021text2event} employ sequence-to-sequence models to extract arguments efficiently. Furthermore, accompanied by sequence-to-sequence models, prompt-tuning methods have also emerged. Recent works on DEAE~\cite{lin2022cup,ma2022prompt,zeng2022ea2e} explore the utilization of various prompts to guide PLMs in extracting arguments.

Up to now, these studies have proposed some solutions to DEAE tasks at different levels, but they rarely consider the entity mentions' relevance directly. Under the latest paradigm of prompt-tuning with generative PLMs, they have not considered the explicit interaction between prompts and documents.

\subsection{DEAE Based on Graph Model}

Graph model is a crucial kind of methods in information extraction, particularly in recent years, where it evaluates documents by constructing various graphs on DEAE tasks. ~\citet{zheng2019doc2edag} first introduced an entity directed acyclic graph to efficiently address DEE. ~\citet{xu2021document} implemented cross-entity and cross-sentence information exchange by constructing heterogeneous graphs. ~\citet{xu2022two} constructed  abstract meaning representation~\cite{banarescu2013abstract} semantic graphs to manage long distance dependencies between trigger and arguments across sentences.

However, these methods based on graph model simply transforms the document into graph structures and then utilize a classification model to assign specific roles to entity mentions. This paradigm make no use of PLMs and is inefficient in extracting all arguments for a given event simultaneously.

Limiting the consideration to just the sequence model or solely the graph model is incomplete. Our research motivation lies in the organic fusion of these two approaches, enabling our method to harness the strengths of both the latest sequence model and graph model. 

\section{ Methodology}

This section begins by introducing the task formulation. We formulate DEAE task as a prompt-based span extraction problem. Given an input instance $\left(X,t,e,R^{(e)}\right)$, where $X = \left\{x_{1},x_{2},...,x_{n}\right\}$ denotes the document, $t\subseteq X$ denotes the trigger word, $e$ denotes the event type and $R^{(e)}$ denotes the set of event-specific role types, we aim to extract a set of spans $A$ as the output. Each $a^{(r)}\in A$ is a segmentation of $X$ and represents an argument corresponding to $r\in R^{(e)}$.

GAM leverages the relations among entity mentions and mask mentions to enhance PLMs for event argument extraction. Our model, depicted in Figure~\ref{fig_model}, comprises three key components: a) semantic mention graph construction from the context, consisting of co-existence, co-reference and co-type relations; b) ensembled graph transformer module for handling the dependencies and interactions in the graph; c) graph-augmented encoder-decoder module with PLMs for argument generation. Subsequent sections will outline our task formulation and elaborate on each component in detail. 


\subsection{Semantic Mention Graph Construction} \label{3.2}
One crucial problem in extracting arguments from the document is mitigating the relevance among entity mentions, as well as the relevance between entity mentions and mask mentions, by capturing co-existence, co-reference and co-type information. Therefore, we introduce a graph construction module that adopts the semantic mention graph to provide a robust semantic structure. This approach facilitates interactions among entity mentions and mask mentions, offering logical meanings of the document from a linguistically-driven perspective to enhance language understanding.

Primarily, as demonstrated in the section \ref{1}, GAM generates an unfilled prompt $p$ with argument placeholders.
GAM initially concatenates the document $X$ with corresponding prompt $p$ respectively to form the input sequences. In DEAE tasks, all extracted arguments should originate from entity mentions in the document. In the prompt-tuning paradigm, the extracted arguments are ultimately filled with placeholders, represented by mask mentions in the prompt. Consequently, we treat all entity mentions mask mentions as nodes in the semantic mention graph. In this module, GAM constructs the semantic mention graph from three perspectives by extracting three types of relations, including co-existence relation within entity mentions and mask mentions, co-reference relation between entity mentions and the co-type relation between mask mentions and entity mentions.

\subsubsection{Co-existence Relation}

In the co-existence relation, GAM focuses on mentions within the same sentence. Intuitively, entity mentions in the same sentence represent all specific information and they are more likely to become arguments for the same event. Mask mentions also represent the same event. The aggregation of the co-existence relation within the mask mentions enables the subsequent sub-modules to better understand which argument roles are present in the current event, thus better reflecting the complete event ontology information in the graph. Therefore, we construct the co-existence relation to enhance the same sentence connection.

If nodes $m_{i}$ and $m_{j}$ are in the same sentence, we establish a direct connection between mentions $m_{i}$ and $m_{j}$. These connections confined within a single sentence in the document or prompt. This relation is reflected in the adjacent matrix $\mathbf{M}_{ex} \in \mathbb{R}^{K\times K}$ of the co-existence relation, where $\mathbf{M}_{ex}[m_{i},m_{j}] = 1$, where $K$ is the total number of the nodes, i.e., the sum of the entity mentions and mask mentions.

Consider Fig.~\ref{fig_dataset} for example, in the same sentence, entity mentions \textit{a homemade bomb} and \textit{Aaron Driver} have an edge connecting them. Similarly, mask mentions \textit{attacker} $\left \langle arg1 \right \rangle$ and \textit{explosiveDevice} $\left \langle arg2 \right \rangle$ also share a direct connection within the same sentence.

\subsubsection{Co-reference Relation}

The co-reference relation aims to make better use of co-reference information between entity mentions. As introduced in the section \ref{1}, it is evident that the co-reference commonly exists in the entire document, showcasing a significant characteristic of co-reference. Hence, we focus on constructing a co-reference relation that captures co-reference relations among entity mentions throughout the entire document.

While, the number of the nodes $K$ is the same as the count of mentions with the co-existence relation. Following the extraction of co-reference relation using the tool \textit{fastcoref}\cite{otmazgin2022f}, we establish direct connection between co-reference entity mentions $m_{k}$ and $m_{l}$. Notably, these connections can occur within sentences or across sentences in the document. Such linkage is represented in the adjacent matrix $\mathbf{M}_{ref} \in \mathbb{R}^{K\times K}$ of the co-reference relation as $\mathbf{M}_{ref}[m_{k},m_{l}] = 1$.

Note that in Fig.~\ref{fig_dataset}, the co-reference entity mentions \textit{Aaron Driver}, \textit{a Canadian man}, \textit{Harun Abdurahman} and \textit{the driver}. There is a direct connection between each of them respectively.

\subsubsection{Co-type Relation}

The co-type relation comprises entity mentions and mask mentions, detailing the relation between the two. Unlike previous methods, we consider the explicit connection between entity mentions and mask mentions in our approach.

A fundamental and logical assumption is that each mask mention should be filled with the same type of entity mentions. In other words, each mask mention should be associated with the same type of entity mentions. Consequently, we compose the third relation to establish co-type connections between mask mentions and entity mentions.

For consistency, the number of the nodes, denoted as $K$, aligns with the count in the previous two relations. Directed connections can be established between mask mention $m_{s}$ and entity mention $m_{t}$ of the same type. These connections link entity mentions in the document to mask mentions in the prompt. These relations are represented in the adjacent matrix $\mathbf{M}_{typ} \in \mathbb{R}^{K\times K}$ of the co-type relation, where $\mathbf{M}_{typ}[m_{s},m_{t}] = 1$. 

As depicted in Fig.~\ref{fig_dataset}, the mask mentions \textit{attacker} $\left \langle arg1 \right \rangle$ and entity mention \textit{Aaron Driver} share the same type, GAM establishes a connection between the two.

\subsection{Ensembled Graph Transformer} \label{3.3}
Several studies ~\cite{zhang2020graph, dwivedi2020generalization} have highlighted drawbacks in graph neural network, including the problem of over-smoothing ~\cite{li2018deeper}. Consequently, we have incorporated the individual approach of graph transformer ~\cite{ying2021transformers, cai2020graph}. Following the extraction of the three types of relations, we utilize ensembled graph transformer structures~\cite{xu2022logiformer} to handle them collectively.

\begin{figure}
\begin{center}
  \includegraphics[scale=0.5]{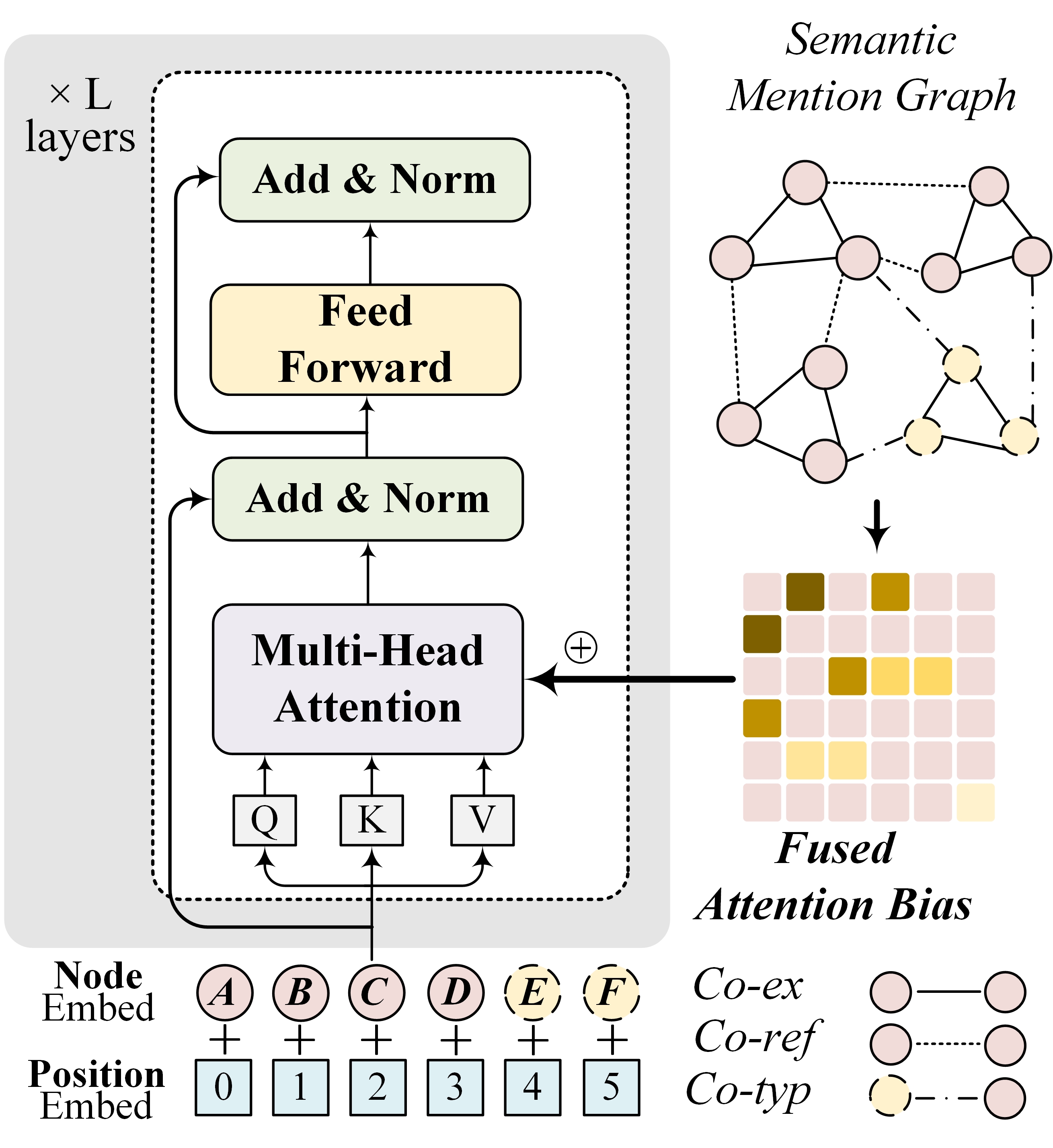}
  \caption{The illustration of graph transformer. The inputs are the node sequence as well as the node position and the outputs are omitted. The Co-ex, Co-ref and Co-typ semantic mention relations are fused as a attention bias. }
  \label{transformer}
\end{center}
\vspace{-0.4cm}
\end{figure}

The merged graph transformer is visually represented in Fig.~\ref{transformer} for a concise overview. First of all, We define text markers as $\mathbf{\left\langle {tgr} \right \rangle / \left \langle /tgr \right \rangle}$ and insert them into the document $X$ before and after the trigger word, respectively. It is essential to obtain the original feature embedding for each node. Given the concatenated sequence of the $i^{th}$ document:

\begin{equation}
	\Tilde{x_{i}} = [x_{1},x_{2},...,\left \langle \mathbf{tgr} \right \rangle,x_{tgr},\left \langle \mathbf{/tgr} \right \rangle,...,x_{n}],
\end{equation} 
where $x_{j}$ represents the $j^{th}$ token in the document, and $tgr$ denotes the index of the trigger word. The document $\Tilde{x_{i}}$ is then encapsulated, together with a prompt template $p$, using a function denoted as $\lambda(\cdot,\cdot)$:

\vspace{-0.3cm}
\begin{equation}
	X_{p} = \lambda(p,\Tilde{x_{i}}) = [C L S] p [S E P] \Tilde{x_{i}} [S E P],
\end{equation}
where $[C L S]$ and $[S E P]$ serve as separators in BART, $X_{p}$ denotes the concatenated input sequence of prompt $p$ and the document $\Tilde{x_{i}}$.

We utilize the BART model as the encoder for obtaining the token-level representation  $\mathbf{V}_{t} \in \mathbb{R}^{N \times d}$ of $X_{p}$, where $N$ is the token numbers of the input sequence and $d$ is the dimension of the hidden state. Subsequently, we extract the order of entity mentions and mask mentions from $\mathbf{V}_{t}$. To obtain the embedding of node $m_{k}$ with the length $L$, we average the token embedding constituting the node:

\begin{equation}
	\mathbf{v}_{k} = \frac{1}{L}\sum\limits_{i = 1}^L {\mathbf{v}_i^{(k)}}.
\end{equation}

We integrate positional embedding and node embedding to maintain the consistency of node order within the document:
\vspace{-0.2cm}
\begin{equation}
	\mathbf{V_{i}} = \mathbf{V_{token}} + Position(\mathbf{V_{token}}),
\end{equation}
where $\mathbf{V_{token}} = [\mathbf{v}_{1}; \mathbf{v}_{2}; ... ; \mathbf{v}_K]$ and  $\mathbf{V_{token}} \in \mathbb{R}^{K\times d}$. The function $Position(\cdot)$ generates a d-dimensional embedding for each node within the input sequence.

This module revolves around multi-head attention mechanism. Firstly, to incorporate graph information into the transformer architecture, we first obtain the fused topology information $M$. Considering the attention bias $\mathbf{M}_{ex}, \; \mathbf{M}_{ref} \; and \; \mathbf{M}_{typ} \in \mathbb{R}^{K\times K}$, these three biases, although having the same dimension, may not contribute equally to the final prediction. To aggregate them effectively, GAM assigns proper hyper-parameters to balance their influence. The representation of $M$ is as follows:

\vspace{-0.5cm}
\begin{equation}
	 \mathbf{M} = \alpha\mathbf{M}_{ex} + \beta\mathbf{M}_{ref} + (1-\alpha-\beta)\mathbf{M}_{typ}.
\end{equation}

Hence, GAM employs the obtained matrix $M$ as attention bias to adjust the self attention formula:

\vspace{-0.2cm}
\begin{equation}
	Att(Q, K, V)^{'} = {\rm softmax}(\frac{Q K^{\rm T}}{\sqrt{d_{k}}} + \mathbf{M}) \cdot V.
 \label{formula}
\end{equation}
where  matrices $Q, K, V \in \mathbb{R}^{K \times d_{k}}$ is the projection of $\mathbf{V_{i}}$ by projection matrices ${\rm W^{Q}, W^{K}, W^{V}} \in \mathbb{R}^{d \times d_{k}}$.

To learn diverse feature representations and improve the adaptability of the graph, we implement multi-head attention mechanism with a specified number of heads, denoted as $MH$:

\vspace{-0.2cm}
\begin{equation}
	MH(Q, K, V) = [Head_{1}; ...; Head_{H}] \cdot {\rm W^{O}},
\end{equation}
where ${\rm W^{O}}\in \mathbb{R}^{(H*d_{k}) \times d_{k}}$ is the linear projection matrix, $Head_{i}=Att_{i}(Q, K, V)^{'}$.

To better capture the diversity and complexity of the attention module, we fuse the last two hidden layers as the updated node features:

\begin{equation}
	\mathbf{V}_{men} = 0.5\cdot\mathbf{V}^{(L-1)} + 0.5\cdot\mathbf{V}^{(L)},
\end{equation}
where $\mathbf{V}_{men} \in \mathbb{R}^{K\times d}$, and $\mathbf{V}^{(L-1)}, \;\mathbf{V}^{(L)} \in \mathbb{R}^{K\times d}$ denote the hidden states of the last two layers.

\subsection{Graph-Augmented Encoder-Decoder Model}

As the previous methods on DEAE ~\cite{lin2022cup,ma2022prompt,zeng2022ea2e} adopt, we choose and expand pre-trained language model BART as our encoder-decoder model.

\begin{table}[]
\begin{tabular}{ccccc}
\toprule
DataSet    & Split & Doc  & Event & Argument \\ \hline
RAMS       & Train & 3,194 & 7,394  & 17,026    \\
           & Dev   & 399  & 924   & 2,188     \\
           & Test  & 400  & 871   & 2,023     \\ \hline
WikiEvents & Train & 206  & 3,241  & 4,542     \\
           & Dev   & 20   & 345   & 428      \\
           & Test  & 20   & 365   & 556      \\ 
\bottomrule
\end{tabular}
\caption{Data statistics of RAMS and WikiEvents.}
\label{dataset}
\end{table}

We have obtained the token-level representation $\mathbf{V}_{t}$ and the updated mention node representation $\mathbf{V}_{men}$. To maintain dimension consistency, we broadcast the feature of each node to all the tokens it encompasses. The transformed features are denoted as $\mathbf{V}_{t}$, $\mathbf{V}_{men}^{'} \in \mathbb{R}^{N \times d}$.

To enhance the ability to perceive semantic mentions, we integrate node embedding into the initial embedding. GAM then configures a proper weight to balance these two features. The resulting fused input embedding $\mathbf{V}$ is as follows:

\begin{equation}
	\mathbf{V} = LN(\mathbf{V}_{t} + \lambda \cdot \mathbf{V}_{men}^{'}),
 \label{lambda}
\end{equation}
where $LN(\cdot)$ denotes the layer normalization operation. Then $\mathbf{V}$ as the input embedding is fed into BART.

To further enhance the effectiveness, GAM incorporates a graph-augmented encoder section of BART. GAM employs the fused topology information $\mathbf{M}$ as attention bias, similar to the graph transformer module. The representation of the self attention formula is adjusted as Eq. \ref{formula}.

For each instance, the graph-augmented BART module can be employed to generate a completed template, replacing the placeholder tokens with the extracted arguments. The model parameter $\theta$ is trained by minimizing the argument extraction loss, which is the conditional probability computed over all instances:

\begin{equation}
	\mathcal L = - \sum\log_{p_{\theta}}{\left( y|X,t,p \right)}.
\end{equation}

\section{Experiments}
\label{sec:append-how-prod}
\subsection{Datasets and baselines}

We conduct comprehensive experiments on two widely recognized DEAE benchmark datasets: RAMS~\cite{ebner2020multi} and WikiEvents~\cite{li2021document}, which have been extensively utilized in previous studies ~\cite{lin2022cup,ma2022prompt,zeng2022ea2e}. As shown in table \ref{dataset}, the RAMS dataset comprises 3,993 paragraphs, annotated with 139 event types and 65 argument roles. The WikiEvents dataset consists of 246 documents, annotated with 50 event types and 59 argument roles.

\begin{table*}[]
\resizebox{\linewidth}{!}{
\begin{tabular}{@{}lcccccccccccc@{}}
\toprule
\multicolumn{1}{c}{\multirow{3}{*}{\textbf{Model}}} & \multicolumn{6}{c}{\textbf{Argument Identification}} & \multicolumn{6}{c}{\textbf{Argument Classification}} \\
\multicolumn{1}{c}{} & \multicolumn{3}{c}{Head Match} & \multicolumn{3}{c}{Coref Match} & \multicolumn{3}{c}{Head Match} & \multicolumn{3}{c}{Coref Match} \\
\multicolumn{1}{c}{} & P & R & F1 & P & R & F1 & P & R & F1 & P & R & F1 \\ \midrule
BERT-CRF & 72.66 & 53.82 & 61.84 & 74.58 & 55.24 & 63.47 & 61.87 & 45.83 & 52.65 & 63.79 & 47.25 & 54.29 \\
ONEIE & 68.16 & 56.66 & 61.88 & 70.09 & 58.26 & 63.63 & 63.46 & 52.75 & 57.61 & 65.17 & 54.17 & 59.17 \\
BART-Gen & 70.43 & 71.94 & 71.18 & 71.83 & 73.36 & 72.58 & 65.39 & 66.79 & 66.08 & 66.78 & \underline{68.21} & 67.49 \\
$\mathrm{\mathbf{EA^{2}E}}$ & \underline{76.51} & \underline{72.82} & \underline{74.62} & \underline{77.69} & \underline{73.95} & \underline{75.77} & \underline{70.35} & \underline{66.96} & \underline{68.61} & \underline{71.47} & 68.03 & \underline{69.7} \\ \midrule
\textbf{GAM} & \textbf{79.05} & \textbf{72.97} & \textbf{75.89} & \textbf{80.36} & \textbf{74.08} & \textbf{77.09} & \textbf{73.47} & \textbf{67.07} & \textbf{70.12} & \textbf{74.59} & \textbf{68.96} & \textbf{71.66} \\
GAM w/o co-ex & 78.34 & 71.66 & 74.85 & 80.28 & 73.09 & 76.52 & 72.86 & 66.80 & 69.70 & 73.69 & 67.29 & 70.34 \\
GAM w/o co-ref & 75.63 & 70.24 & 72.84 & 76.07 & 70.72 & 73.30 & 69.85 & 64.05 & 66.82 & 70.53 & 64.74 & 67.51 \\
GAM w/o co-typ & 76.44 & 70.95 & 73.59 & 78.62 & 72.34 & 75.35 & 71.96 & 67.16 & 69.48 & 72.81 & 66.46 & 69.49 \\
GAM w/o G.T. & 78.46 & 70.52 & 74.28 & 79.45 & 71.4 & 75.21 & 71.34 & 64.12 & 67.54 & 72.33 & 65.01 & 68.48 \\
GAM w/o N.E. & 77.08 & 72.29 & 74.61 & 78.03 & 73.18 & 75.53 & 70.64 & 66.25 & 68.38 & 71.59 & 67.14 & 69.29 \\
GAM w/o bias & 76.85 & 70.16 & 73.35 & 77.82 & 71.05 & 74.28 & 70.23 & 64.12 & 67.04 & 71.21 & 65.01 & 67.97 \\ \bottomrule
\end{tabular}
}
\caption{Overall performance on WikiEvents dataset. In the results, the best-performing model is highlighted, and the second best is underlined. \textbf{G.T.}: graph transformer module. \textbf{N.E.}: node embedding module. \textbf{bias}: attention bias for graph transformer and BART encoder module.}
\label{results1}
\end{table*}

We deem an argument span as correctly identified when its offsets align with any of the reference arguments of the current event (i.e., \textbf{Argument Identification}), and as correctly classified when its role matches (i.e., \textbf{Argument Classification}). Furthermore, we evaluate the argument extraction performance using Head Match F1 and Coref Match F1 metrics on the WikiEvents dataset, where Head Match indicates alignment with the head of the span, and Coref Match indicates an exact match of the span with all co-reference spans. In the case of the latter,  full credit is assigned when the extracted argument is coreferential with the gold-standard argument.

We compare GAM with several state-of-the-art models in two categories: (1) \textbf{FEAE}~\cite{wei2021trigger}, \textbf{EEQA}~\cite{du2020event}, \textbf{BART-Gen}~\cite{li2021document}, \textbf{PAIE}~\cite{ma2022prompt} on RAMS dataset; (2) \textbf{BERT-CRF}~\cite{shi2019simple}, \textbf{ONEIE}~\cite{lin2020joint}, \textbf{BART-Gen}~\cite{li2021document}, $\mathrm{\mathbf{EA^{2}E}}$~\cite{zeng2022ea2e} on WikiEvents dataset. Among them, \textbf{BERT-CRF} is a semantic role labeling method, \textbf{ONEIE} is a graph-based method, \textbf{FEAE} and \textbf{EEQA} utilize QA patterns, whereas \textbf{BART-Gen}, \textbf{PAIE}, and $\mathrm{\mathbf{EA^{2}E}}$ employ different prompts directly.

\subsection{Implementation Details}

\begin{table}[]
\resizebox{\linewidth}{!}{
\begin{tabular}{@{}lcc@{}}
\toprule
\multicolumn{1}{c}{\textbf{Model}} & \textbf{Argument Identification} & \textbf{Argument Classification} \\ \midrule
FEAE & 53.5 & 47.4 \\
EEQA & 48.7 & 46.7 \\
BART-Gen & 51.2 & 47.1 \\
EEQA-BART & 51.7 & 48.7 \\
PAIE & \underline{55.6} & \underline{53.0} \\ \midrule
\textbf{GAM} & \textbf{56.83} & \textbf{54.20} \\
GAM w/o co-ex & 54.52 & 52.19 \\
GAM w/o co-ref & 52.86 & 50.65 \\
GAM w/o co-typ & 53.16 & 51.82 \\
GAM w/o G.T. & 54.24 & 53.02 \\
GAM w/o N.E. & 53.64 & 51.17 \\
GAM w/o bias & 53.94 & 52.45 \\ \bottomrule
\end{tabular}
}
\caption{Overall performance on RAMS dataset.}
\label{results2}
\vspace{-0.2cm}
\end{table}

GAM extends upon the BART-style encoder-decoder transformer structure. Each model, including baselines and GAM, is trained for 4 epochs with a batch size of 4, utilizing NVIDIA-V100 with 32GB DRAM. The model is optimized using the Adam optimizer with a learning rate of 3e-5, $\alpha=0.3$, $\beta=0.4$ and $\lambda = 0.015$. These hyper-parameters are meticulously selected through grid search, based on model's performance on the development set.\footnote{The learning rate is chosen in \{3e-5, 5e-5\}, $\alpha$ and $\beta$ is chosen from \{0.1, 0.2, 0.3, 0.4, 0.6, 0.7, 0.8\}, and $\lambda$ is chosen from \{0.01, 0.015, 0.02, 0.03, 0.04, 0.05\}.}

\subsection{Comparison Results}

Tables \ref{results1} and \ref{results2} demonstrate the superior performance of our proposed GAM compared to strong baseline methods across various datasets and evaluation metrics. Specifically, on the WikiEvents dataset, our model achieves a notable 1.32\% improvement in absolute argument identification F1 and a 1.96\% improvement in argument classification F1. Similarly, on the RAMS dataset, GAM exhibits the improvements with a 1.23\% increase in argument identification and 1.20\% in argument classification F1 scores. These results underscore the outstanding performance of our proposed method.

Furthermore, our graph-augmented encoder-decoder model outperforms graph-based methods and directly prompt-tuning encoder-decoder methods, including ONEIE and BART-Gen. From the experimental results shown in Tables \ref{results1} and \ref{results2}, we can conclude that: (1) Compared to graph-based methods, GAM can utilize rich information from the graph to enhance the initial embedding and the encoder part of encoder-decoder model. (2) Compared to directly prompt-tuning encoder-decoder methods, These results emphasize the effectiveness of our semantic mention graphs in leveraging the BART architecture, enhancing semantic interactions within documents, and bridging the gap between documents and prompts.

\subsection{Ablation Studies}

\begin{figure}
\begin{center}
  \includegraphics[scale=0.4]{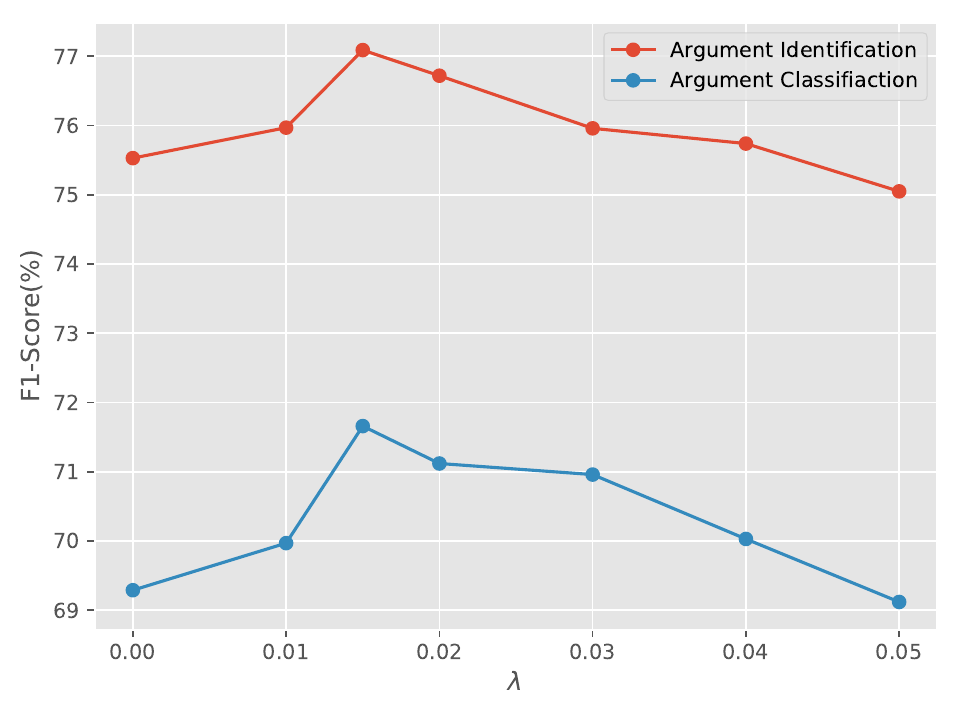}
  \vspace{-0.2cm}
  \caption{The illustration of ablation study on WikiEvents dataset. The model performances under different $\lambda$.}
  \label{ablation1}
  \vspace{-0.6cm}
\end{center}
\end{figure}

\begin{figure}
\begin{center}
  \includegraphics[scale=0.4]{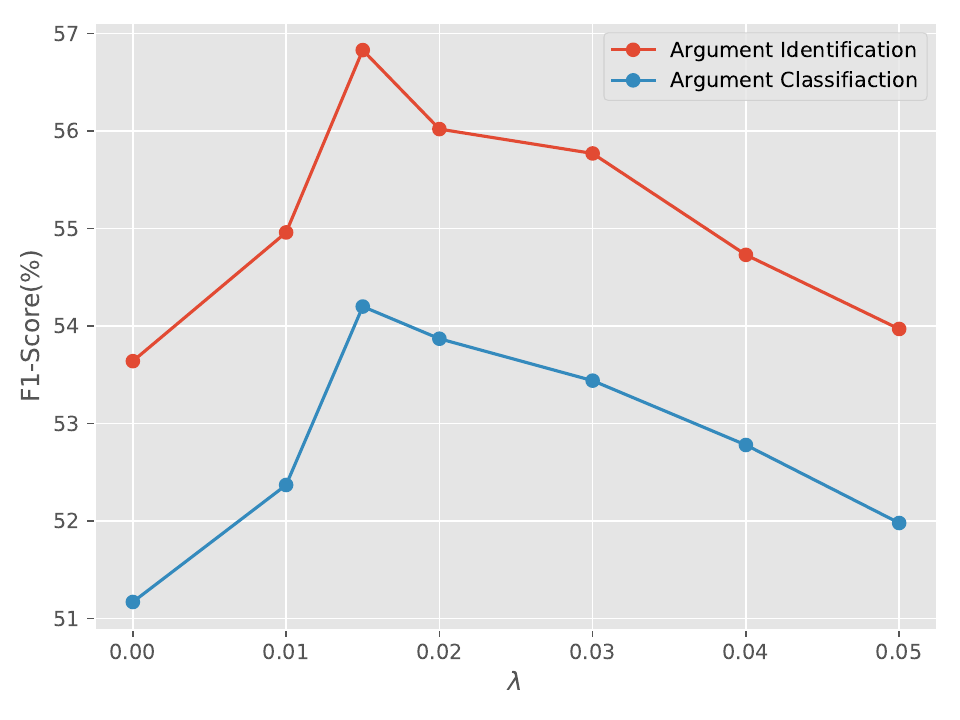}
  \vspace{-0.2cm}
  \caption{The different performance on RAMS dataset under different $\lambda$.}
  \vspace{-0.7cm}
  \label{ablation2}
\end{center}
\end{figure}

\begin{figure*}[]
	\large
	\centering
	\includegraphics[scale=0.5]{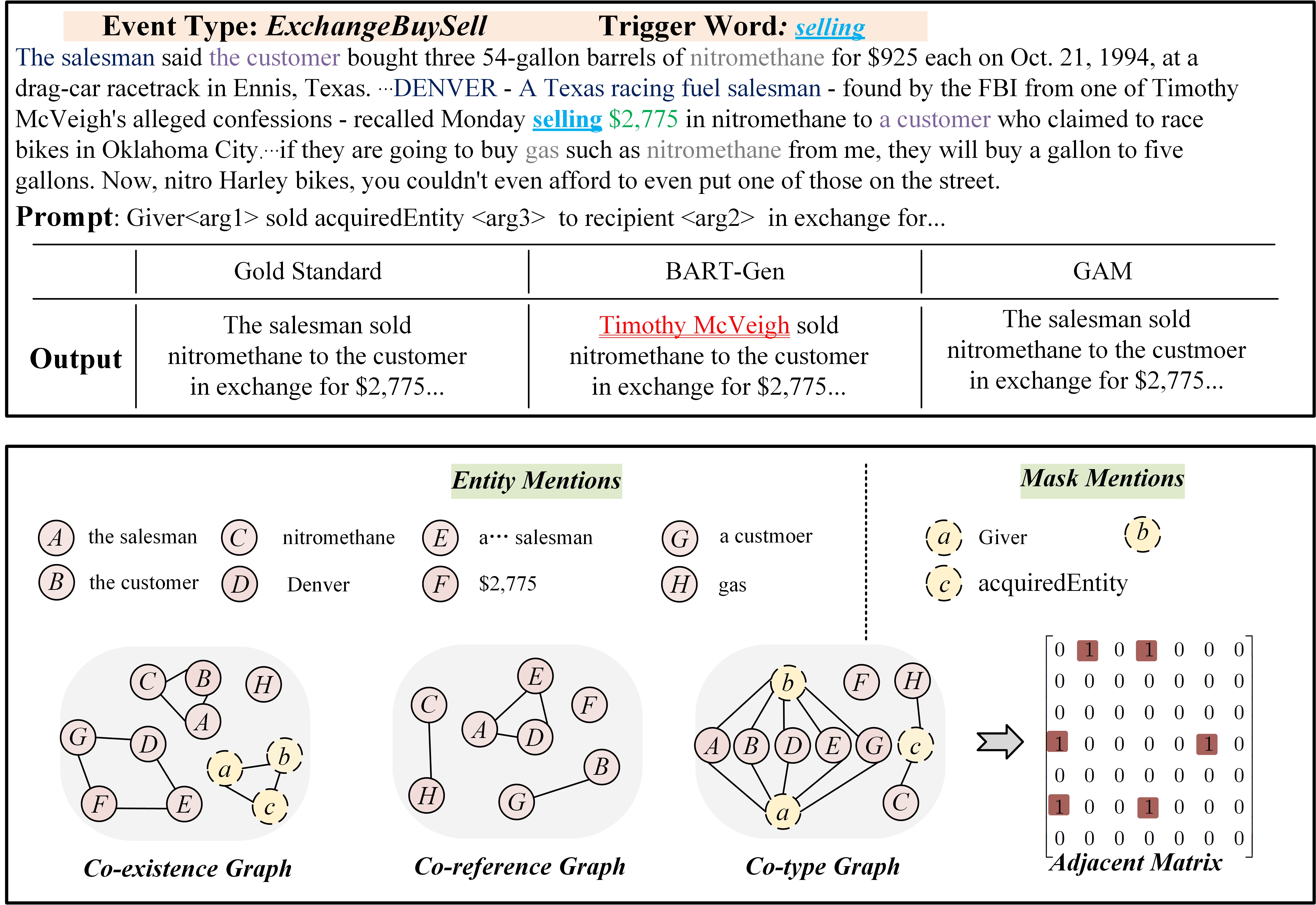}
	\caption{The illustration of an DEAE case. This case mainly showcases the construction of semantic mention graph and compares the extraction results between BART-Gen and GAM.}
 \vspace{-0.2cm}
	\label{casestudy}
	\vspace{-0.1cm}
\end{figure*}

In this section, we assess the effectiveness of our primary components by systematically removing each module one at a time. The components are as following: (1) three types of relations. Here, we analyze the gain brought by these relations by considering the removal of one of the three; (2) graph transformer. We exclude the graph transformer, thereby disregarding the updating of node representation in the semantic mention graph; (3) node embedding. We eliminate the node embedding component from the input of the BART encoder-decoder module, retaining only the initial embedding; (4) attention bias. We withhold the attention bias from both the graph transformer and the BART encoder module.

The results of ablation studies are summarized in Table \ref{results1} and Table \ref{results2}. We can observe that all of the three types of relations, graph transformer, node embedding and attention bias modules can help boost the performance of DEAE.

Regarding the module of the semantic mention graph construction, we remove one of the three relations at a time to observe the decrease it causes. According to the ablation results, the co-reference relation contributes the most among the three types of relations, followed by co-type relation and co-existence relation. It is evident that the co-reference relation significantly reduces ambiguity, enhances the accuracy of DEAE, and provides a more comprehensive, consistent, and precise semantic representation. The decrease brought by the co-type relation follows closely behind because, ideally, the correctly extracted arguments should all come from corresponding co-type entity mentions in the original document.

Moreover, the absence of the graph transformer module leads to a obvious drop in performance, with F1 score decreasing by more than 2 points on both RAMS and WikiEvents datasets. This clearly emphasizes the crucial role of the graph transformer in updating nodes. Similar patterns are observed in other modules, underscoring the effectiveness of each component in enhancing argument extraction. We are pleasantly surprised to discover that withholding attention bias from both the graph transformer and the BART encoder module resulted in the largest decrease, excluding the semantic mention graph construction module. This is because, in the transformer architecture, the attention mechanism tends to allocate more attention to the emphasized parts.

In particular, Dropping out all of the above modules—essentially eliminating all components related to the graph—results in the variant model regressing to BART-Gen. BART-Gen is a standard model that relies solely on prompts and PLMs. Upon reviewing the results in Table \ref{results1} and Table \ref{results2}, GAM outperforms BART-Gen by 4.17\% on the WikiEvents dataset and 5.5\% on the RAMS dataset. This comparison strongly emphasizes the significant performance enhancement achieved by the graph-enhanced model over BART.

\subsection{Supplementary Analysis}

Throughout the experiments, hyper-parameters are employed in many places. Due to space constraints, we focus on analyzing one specific parameter—namely, the node embedding weight $\lambda$ used to consolidate the initial embedding fed into BART. As shown in Eq. \ref{lambda},  $\mathbf{V}_{t}$  refers to the initial embedding of the document, while $\mathbf{V}_{men}^{'}$ refers to the updated node embedding, reflecting the GAM model's modeling of nodes in the semantic mention graph, addressing the independent modeling of entity mentions and document-prompt isolation. $\lambda$ is the balanced weight of $\mathbf{V}_{men}^{'}$, enhancing the input of the graph-augmented encoder-decoder module. The results corresponding to different values of $\lambda$ are presented in Fig.~\ref{ablation1} and Fig.~\ref{ablation2}.

The results demonstrate that the optimal performance is achieved when $\lambda$ is set to 0.015. A decrease in the hyper-parameter $\lambda$ implies less consideration of node features and underutilization of semantic information. Conversely, as $\lambda$ increases, additional semantic information is incorporated into the initial embedding. However, this might be detrimental to subsequent decoder stages because the encoder-decoder architecture heavily depends on the transmission of initial embedding within this context.

\section{Case Study}
 Fig.~\ref{casestudy} presents a representative example from the WikiEvents dataset, illustrating the process of graph-augmented DEAE. Initially, the graph construction module comprises nodes representing all entity mentions and mask mentions, along with edges depicting three semantic mention relations. In this instance, nodes are represented as circles in green and gray. GAM generates the semantic mention graph based on these relations. The connections efficiently capture the co-existence, co-reference and co-type information within and between the document and the prompt, highlighting GAM's interpretability capability. 
 
Finally, GAM accurately extracts arguments corresponding to their respective roles using an unfilled prompt $p$. As depicted in Fig.~\ref{casestudy}, the output of BART-Gen differs from that of GAM. When compared to the gold standard, BART-Gen incorrectly identifies the argument role \textit{Giver} due to its failure in considering the three types of relations within and between the document and the prompt. Conversely, GAM accurately aligns with the gold standard.

While effective, GAM can inadvertently propagate errors during graph construction. Furthermore, a scenario might arise where an argument role lacks a corresponding argument in the document. In such cases, the co-type relation may still assign edges of the same type of entity mention to these mask mention nodes.

\section{Conclusion}
We propose an end-to-end framework named semantic mention Graph Augmented Model to address the independent modeling of entity mentions and the document-prompt isolation problems. Firstly, GAM constructs a semantic mention graph by creating three types of relations: co-existence, co-reference and co-type relations within and between mask mentions and entity mentions. Secondly, The ensembled graph transformer module is utilized to handle the mentions and their three semantic relations. Lastly, the graph-augmented encoder-decoder module integrates the relation-specific graph into the input embedding and optimize the encoder section with topology information to enhance the performance of PLMs. Extensive experiments report that GAM achieves the new state-of-the-art performance on two benchmarks. 

In the future, we plan to delve into DEAE within the framework of Large Language Models (LLM)~\cite{xu2023symbol}. Due to the ambiguity~\cite{liu2023we} and polysemy~\cite{laba2023contextual} inherent in entity mentions within documents, LLM faces limitations in DEAE. We aim to leverage the semantic mention graph to provide guidance to LLM in DEAE. 
Furthermore, we will strive to integrate prior knowledge and employ logical reasoning~\cite{DBLP:conf/sigir/LinLXPZZZ22,DBLP:conf/acl/LinL0XC23} to enhance event extraction with greater precision and interpretability.

\section*{Acknowledgement}
This work was supported by National Key Research and Development Program of China (2022YFC3303600), National Natural Science
Foundation of China (62137002, 62293553, 62176207, 62192781, 62277042 and 62250009), "LENOVO-XJTU" Intelligent Industry Joint Laboratory Project, Natural Science Basic Research Program of Shaanxi (2023-JC-YB-593), the Youth Innovation Team of Shaanxi Universities, XJTU Teaching Reform Research Project "Acquisition Learning Based on Knowledge Forest", Shaanxi Undergraduate and Higher Education Teaching Reform Research Program(Program No.23BY195).

\nocite{*}
\section*{Bibliographical References}\label{sec:reference}

\bibliographystyle{lrec-coling2024-natbib}
\bibliography{lrec-coling2024-example}


\end{document}